\documentclass[sigconf,natbib=true,anonymous=false]{acmart}
\usepackage{amsmath}
\usepackage{multirow}
\usepackage{subfigure}

\AtBeginDocument{%
  \providecommand\BibTeX{{%
    \normalfont B\kern-0.5em{\scshape i\kern-0.25em b}\kern-0.8em\TeX}}}

\setcopyright{acmcopyright}
\copyrightyear{2018}
\acmYear{2018}
\acmDOI{10.1145/1122445.1122456}

\acmConference[Woodstock '18]{Woodstock '18: ACM Symposium on Neural
  Gaze Detection}{June 03--05, 2018}{Woodstock, NY}
\acmBooktitle{Woodstock '18: ACM Symposium on Neural Gaze Detection,
  June 03--05, 2018, Woodstock, NY}
\acmPrice{15.00}
\acmISBN{978-1-4503-XXXX-X/18/06}

\begin{document}

\title{Deep Knowledge Tracing with Learning Curves}

\author{Shanghui Yang}
\email{zqws1018@outlook.com}
\affiliation{%
  \institution{East China Normal University}
  \city{Shanghai}
  \country{China}
}

\author{Mengxia Zhu}
\email{51195100046@stu.ecnu.edu.cn}
\affiliation{%
  \institution{East China Normal University}
  \city{Shanghai}
  \country{China}
}
  
\author{Xuesong Lu}
\email{xslu@dase.ecnu.edu.cn}
\affiliation{%
  \institution{East China Normal University}
  \city{Shanghai}
  \country{China}
}


\begin{abstract}
  Knowledge tracing (KT) has recently been an active research area of computational pedagogy. The task is to model students' mastery level of knowledge concepts based on their responses to the questions in the past, as well as predict the probabilities that they correctly answer subsequent questions in the future. KT tasks were historically solved using statistical modeling methods such as Bayesian inference and factor analysis, but recent advances in deep learning have led to the successive proposals that leverage deep neural networks, including long short-term memory networks, memory-augmented networks and self-attention networks. While those deep models demonstrate superior performance over the traditional approaches, they all neglect the explicit modeling of the learning curve theory, which generally says that more practice on the same knowledge concept enhances one's mastery level of the concept. Based on this theory, we propose a Convolution-Augmented Knowledge Tracing (CAKT) model in this paper. The model employs three-dimensional convolutional neural networks to explicitly learn a student's recent experience on applying the same knowledge concept with that in the next question, and fuses the learnt feature with the feature representing her overall latent knowledge state obtained using a classic LSTM network. The fused feature is then fed into a second LSTM network to predict the student's response to the next question. Experimental results show that CAKT achieves the new state-of-the-art performance in predicting students' responses compared with existing models. We also conduct extensive sensitivity analysis and ablation study to show the stability of the results and justify the particular architecture of CAKT, respectively. 
  \end{abstract}



\maketitle

\section{Introduction}\label{sec:introduction}
One important aspect in education is to continuously estimate students' mastery level of knowledge, or simply termed knowledge state. According to students' knowledge state, tutors may properly design personalized learning paths and help them master all the knowledge in an efficient manner. Among those alternative estimation methods, \emph{knowledge tracing} (KT) models students' changing knowledge state by tracking their interactions with coursework, i.e., a sequence of questions being solved. By observing whether a student correctly answers a sequence of questions in the history, where each question contains a particular knowledge concept, a KT model adjusts her knowledge state over time and also predicts her performance on the questions in future. Thanks to the ease of interpretation and adoption, the philosophy of knowledge tracing has been widely adopted in intelligent tutoring systems and recently in MOOC platforms~\cite{Kaplan2016Higher, Mcgreal2013Massive}.

Knowledge tracing models are used to be constructed using statistical cognitive modeling methods such as Bayesian inference with a Hidden Markov model~\cite{corbett1994knowledge,yudelson2013individualized} and factor analysis using logistic regressions~\cite{cen2006learning,pavlik2009performance,chi2011instructional}. Recently, researchers have turned to train neural network based models for knowledge tracing due to the availability of massive educational data released by large MOOC platforms and educational institutions. These deep models have shown superior performance over traditional methods in terms of prediction accuracy. In the pioneering work~\cite{piech2015deep}, Piech et al. proposed the DKT model using an LSTM network, which significantly improved the overall AUC of predicting students' responses to questions. The model reads each student-question interaction (consisting of a question and the correctness of the student's answer) sequentially and predicts whether the student answers the next question correctly. Inspired by this work, a series of deep learning models were proposed to target various aspects in the knowledge tracing task, including DKVMN~\cite{zhang2017dynamic}, EKT~\cite{huang2019ekt}, SKVMN~\cite{abdelrahman2019knowledge}, SAKT~\cite{pandey2019self}, AKT~\cite{ghosh2020context} and CKT~\cite{shen2020convolutional}, etc. Readers may refer to Section~\ref{sec:related} for a detailed literature review of existing deep KT models.

Despite the major advances in predicting student performance, we notice that existing deep KT algorithms have almost neglected the explicit modeling of the ``learning curve'', which generally states that more practice brings more improvement on a skill. We show in this work that explicit modeling of the learning curve can indeed boost the prediction performance of deep KT models. Newell and Rosenbloom~\cite{newell1981mechanisms} first theorized the ubiquitous phenomenon and found that the error rate of performance and the amount of practice have a power relationship in diverse learning tasks. They depicted the relationship using the following simple equation:
\begin{equation}
Y=aX^{-b},
\end{equation}
where $Y$ is the error rate or the cost to complete a task, $X$ is the number of previous trials using a skill needed by the task, $a$ is the difficulty of the skill and $b$ is the learning rate of the skill. Other variants of equations pertaining to the learning curve include s-curve and exponential growth functions~\cite{Ritter2002thelearning,leibowitz2010exponential}. Although learning curves may demonstrate different shapes, they all reveal that the more someone practices on a skill, the better she performs on it. Inspired by this phenomenon, Cen et al.~\cite{cen2006learning} proposed the Learning Factors Analysis (LFA) model, where they used as the independent variables the number of opportunities to practice a skill in the past as well as the interaction between the skill and the number of practices. The model was subsequently improved in the PFA~\cite{pavlik2009performance} and IFA~\cite{chi2011instructional} models, where more sophisticated features regarding the learning curve were constructed and the prediction performance was further improved. Surprisingly, we find that this simple idea has not been sufficiently explored in existing deep KT models. In DKT~\cite{piech2015deep} and DKVMN~\cite{zhang2017dynamic}, all the student-question interactions in the past are equally treated, although most of them may require a different knowledge concept with that of the question being answered. In other words, when predicting a student's response to the next question pertaining to a knowledge concept, the models haven't given particular bias to the student's experience of applying the same knowledge concept in the past. In the models such as EERNN\textbackslash EKT~\cite{su2018exercise,huang2019ekt}, SAKT~\cite{pandey2019self} and AKT~\cite{ghosh2020context}, the heterogeneous impacts of the past interactions are modeled using the (soft/self-) attention mechanism. In this way, the experience of practice in the past on the same concept might be implicitly amplified, but the interference of the unrelated questions still exists. The two models SKVMN~\cite{abdelrahman2019knowledge} and DKT+forgetting~\cite{nagatani2019augmenting} have attempted to explicitly model the impact of the past performance on the same knowledge concept. Nevertheless, SKVMN needs to empirically define a threshold for the similarity between two knowledge concepts and then pick past trials on the same concept based on the threshold, whereas DKT+forgetting uses simple handcrafted features to represent repeated practices on the same concept. Therefore both of them rely on empirically-determined settings which may lead to sub-optimal performance.

To bridge this gap, we propose a new deep KT model, which is called \emph{Convolution-Augmented Knowledge Tracing}, or simply CAKT, to explicitly model a student's experience on the same knowledge concept with the one covered by the question to be answered. At the core of the new architecture are three-dimensional convolutional neural networks (3D-ConvNets) for capturing the information from the recent student-question interactions on the same concept. Specifically, when predicting a student's response to the next question at time $t$+$1$ pertaining to a knowledge concept $c$, we fetch $k$ recent interactions of her before $t$+$1$ that also cover the concept $c$, and represent them using vectorized embeddings. Then we reshape the embeddings into matrices and stack them in the chronological order to form a three-dimensional tensor. This tensor represents the student's experience on applying $c$ in the past. We then use a 3D-ConvNets module to extract the latent knowledge state of the student on the concept $c$ from the tensor and denote this latent feature by $\mathbf{m}_t$. In this way, we explicitly model the learning curve by only extracting the feature from past trials on $c$. Apart from the 3D-ConvNets, we leverage the classic LSTM network for the KT problem~\cite{piech2015deep} to extract the latent knowledge state on all the concepts from the student's interactions until $t$. We denote this feature by $\mathbf{h}_t$. To augment the learning curve theory into the model, we borrow the idea of the threshold mechanism in LSTM\textbackslash GRU and propose a fusion gate to fuse the two latent features. The intuition is that we leverage both the overall knowledge state since there might be other concepts relevant to $c$, and the particular knowledge state of $c$ since the past experience of applying $c$ largely affects the response to the next question. Finally, the fused feature is transformed using a second LSTM layer to predict the student's response to the next question at time $t$+$1$.

Here we point out that one may consider to replace the 3D-ConvNets with other networks such as RNNs\textbackslash LSTM~\cite{rumelhart1986learning,hochreiter1997long}, Transformer~\cite{vaswani2017attention} and ordinary CNNs~\cite{lecun1989backpropagation} to capture the latent feature $\mathbf{m}_t$, but in practice they all result in sub-optimal performance as shown in the experimental section. We conjecture the reasons are as follows. When using RNNs\textbackslash LSTM or Transformer, the input student-question interactions are embedded into vectors similarly to DKT~\cite{piech2015deep}, EKT~\cite{huang2019ekt} and SAKT~\cite{pandey2019self}, so that the extracted latent features $\mathbf{m}_t$ and $\mathbf{h}_t$ are very likely to be redundant and thus bring no performance gain. When using CNNs, a common method is to arrange the vectorized embeddings into a matrix (feature map) and apply convolutional operations on it. As such neither the knowledge state contained in each embedding nor the evolution of the knowledge state is well captured because the filters only focus on the local patterns in the feature maps. To overcome the drawbacks, we are inspired by the video analysis tasks~\cite{tran2015learning,tran2018closer}, which use 3D-ConvNets to extract features from a sequence of frames, and propose the aforementioned architecture. On one hand, we reshape the embeddings into matrices so that we could leverage convolutional modules to extract features differently from those extracted by the LSTM network. On the other hand, the 3D-ConvNets use the first two dimensions of the filters to capture the latent knowledge state in each reshaped matrix and the third dimension to capture the evolution of the knowledge state from the tensor. We conduct extensive empirical study in Section~\ref{sec:experiment}, and demonstrate that CAKT outperforms main existing KT models as well as its alternative architectures in predicting students' responses to questions in the KT task. Particularly, the big improvements over DKT prove the importance of modeling the learning curve theory, since the performance gain mainly comes from the 3D-ConvNets module.

The rest of the paper is organized as follows. In Section~\ref{sec:related} we present a comprehensive literature review of existing knowledge tracing models and introduce some basics pertaining to three-dimensional convolutional neural networks. In Section~\ref{sec:input} we formally define the knowledge tracing problem. We describe the architecture details of the proposed CAKT model in Section~\ref{sec:models}, followed by extensive performance evaluation in Section~\ref{sec:experiment}. Finally, we conclude the work in Section~\ref{sec:conclusion}.

\section{Related Work}\label{sec:related}
\subsection{Cognitive Knowledge Tracing Models}
The idea of knowledge tracing is proposed in~\cite{corbett1994knowledge}, where the authors construct a tutoring system with a production rule cognitive model of programming knowledge concepts. As a student solves programming questions, the system estimates the probability that the student has learned each concept, i.e., estimating the student's programming knowledge state. They propose a tracing model called Bayesian Knowledge Tracing (BKT). The model uses four parameters for each knowledge concept and employs a Hidden Markov model with Bayesian inference to fit the sequence data of student-question interactions on the learning system. Later, a series of factor analysis models using logistic regression are proposed~\cite{cen2006learning,pavlik2009performance,chi2011instructional}. These models generalize the learning curve theory~\cite{newell1981mechanisms} and manually construct input features such as the number of attempts for each question, the number of correct and incorrect attempts and the number of mentions for each knowledge concept. The results show these simple models have predictive power comparable to BKT.

\subsection{Knowledge Tracing with Deep Learning}
The Deep Knowledge Tracing (DKT) model~\cite{piech2015deep} first applies deep learning on the KT task. DKT uses an LSTM network~\cite{graves2013speech, sutskever2014sequence} to learn from the student-question interaction sequences. At each step, the LSTM unit takes as input an interaction tuple representing which question is answered and whether the answer is correct. The tuple is encoded using a one-hot vector. The output is a vector of length equal to the number of knowledge concepts, where each element is a probability representing the predicted mastery level of a concept. When predicting a student's response to the next question, the element of the output vector corresponding to the concept covered by the question is used to predict the probability of correctness. Finally, the loss function is computed as the sum of binary cross entropy between the predicted responses and the ground-truth responses. DKT significantly outperforms the cognitive models and its variants in terms of AUC.

Then inspired by memory-augmented neural networks~\cite{graves2016hybrid, santoro2016meta}, Zhang et al.~\cite{zhang2017dynamic} propose Dynamic Key-Value Memory Networks (DKVMN) to improve DKT's structure. The assumption is that the hidden state in the LSTM network has limited power to represent hidden knowledge state, therefore DKVMN introduces memory matrices to store more abundant hidden information. DKVMN uses two memory matrices, where one is static and used to store the latent knowledge concepts of the questions, and the other is dynamic and used to represent student knowledge state. Each matrix slot stores the state of one concept. DKVMN updates the knowledge state of a student by reading from and writing to the dynamic matrix using correlation weights computed from the input question and the static matrix. Following this work, Abdelrahman et al. propose Sequential Key-Value Memory Networks (SKVMN)~\cite{abdelrahman2019knowledge} to capture the dependencies between questions. They assume that the predicted response to the next question only depends on the previous interactions pertaining to the questions with similar concepts. At each step, they introduce an additional hop-LSTM network before the output layer of DKVMN, whose LSTM units connect only the hidden states of the steps pertaining to the dependent questions. To some extent, SKVMN attempts to use the hop-LSTM structure to model the learning curve. However, whether two questions are dependent is determined by an empirically-defined threshold of the similarity between the questions computed using an triangular membership function~\cite{klir1996fuzzy}.

Another line of work attempts to incorporate additional features in model input. For example, EERNN~\cite{su2018exercise} uses a Bi-LSTM network to obtain the text embedding of each question, and concatenates the embedding with that of the corresponding student-question interaction tuple. The concatenated embeddings are fed into a LSTM network sequentially. EERNN employs two different architectures to obtain the hidden state of the past interactions. One uses the hidden state of the last LSTM unit as in DKT. The other uses an attention mechanism to aggregate all the hidden states of the past LSTM units. Experimental results show the attention mechanism brings additional boost on prediction performance. Later, Huang et al.~\cite{huang2019ekt} extend EERNN and propose the EKT model, which borrows the idea of memory networks and replaces the hidden state in the LSTM network with a hidden matrix. The DKT+forgetting~\cite{nagatani2019augmenting} model uses manually-constructed features pertaining to the forgetting behavior in the learning process, and feeds the features as additional information into the DKT model. Two of the constructed features are related to the learning curve theory, which are the time gap to the previous question with the same concept and the number of past attempts on the same concept. However, the handcrafted features are not robust and usually have low discriminative power. Other work of this line includes DKT-DSC~\cite{minn2018deep} that clusters students in every few steps and uses the clustering results as additional input, PDKT-C~\cite{chen2018prerequisite} that incorporates prerequisite relations between knowledge concepts as additional constraints, and CKT~\cite{shen2020convolutional} that constructs the features pertaining to students' historical performance and extracts the personalized learning rate feature using one-dimensional convolutions, etc.

Other work investigates the utility of recently proposed architectures. For example, Pandey et al.~\cite{pandey2019self} propose Self-Attentive Knowledge Tracing (SAKT), with the hope to handle the data sparsity problem by using the Transformer architecture~\cite{vaswani2017attention}. When predicting the response to the next question, SAKT attends to all the previous student-question interactions by assigning a learnable weight to each of them. Ghosh et al.~\cite{ghosh2020context} propose Attentive Knowledge Tracing (AKT), which uses a series of attention networks to draw connections between the next question and every question the learner has responded to. They first use two self-attended encoders to embed questions and knowledge concepts, respectively, and then use the monotonic attention mechanism to retrieve the latent knowledge state at each step. The mechanism down-weights the importance of questions in the distant past to mimic the forgetting behavior. Nakagawa et al.~\cite{nakagawa2019graph} leverage graph neural networks and propose Graph-based Knowledge Tracing (GKT). They construct a graph such that the nodes are the knowledge concepts and there is an edge between two nodes if the corresponding concepts are related. When a student answers a question associated with a particular concept, GKT first aggregates the node features related to the concept, and then updates simultaneously the student’s knowledge state on the concept as well as the related concepts. 

All existing methods have not sufficiently investigated how to incorporate the learning curve theory into deep models. Among them, the attention-augmented and attention-based methods may implicitly model the impact of the past attempts on the same concept, but still fail to eliminate the influence of unrelated questions. SKVMN and DKT+forgetting consider the features related to the learning curve, but the construction of the features relies on empirically-determined settings which may lead to sub-optimal performance.

\subsection{Three-dimensional Convolutional Networks}
Convolutional neural networks have been successfully adopted in the field of computer vision, such as object detection~\cite{redmon2018yolov3}, image segmentation~\cite{he2017mask} and optical character recognition~\cite{shi2016end}. Inspired by this great success, three-dimensional convolutional neural networks~\cite{tran2015learning, tran2018closer} are proposed to handle video analysis tasks. A video can be viewed as a sequence of images (frames) and formally represented using a $\mathrm{D} \times \mathrm{H} \times \mathrm{W}$ tensor, where $\mathrm{D}$ represents the depth or time of the video, $\mathrm{H}$ and $\mathrm{W}$ represent the height and width of each frame, respectively. The filters of 3D-ConvNets have also three dimensions accordingly. Tran et al.~\cite{tran2015learning} demonstrate that 3D-ConvNets can better model the temporal information than two-dimensional convolutional networks. When using two-dimensional convolutions, all the frames are convolved using the same 2D filters and therefore the temporal information is neglected, whereas 3D filters in 3D-ConvNets preserve temporal information during the convolution operations. We therefore leverage this property and use 3D-ConvNets to learn from the sequence of reshaped interaction matrices pertaining to the same knowledge concepts, with the expectation that both the knowledge state at each step and the evolution of the knowledge state are learned. 

\section{Problem Definition}
\label{sec:input}
At each step, knowledge tracing takes as input a sequence of previous student-question interactions and outputs the prediction of the student's response to the next question. Formally, the problem of knowledge tracing can be defined as follows.\\

\textbf{Definition of (deep) knowledge tracing.} \emph{For each student, denote by $q_i$ the $i^{th}$ question she answers and by $a_i$ the corresponding response. At each step $t$, given a sequence of previous student-question interactions $\mathcal{X} = \{x_1, x_2, \dots, x_t\}$, where $x_i=(q_i, a_i)$, knowledge tracing predicts the student's response $a_{t+1}$ to the next question $q_{t+1}$, i.e., the probability $\mathrm{P}(a_{t+1} = 1 | q_{t+1}, \mathcal{X})$ that the student answers the next question correctly.}
\\

In the above definition, $a_i$ is a binary variable where $1$ represents the student's answer is correct and $0$ otherwise, $q_i$ is represented using a one-hot vector $\textbf{e}_i$ with length $M$, where $M$ is the number of distinct questions\textbackslash concepts\footnote{We follow the convention that each question covers exactly one knowledge concept and all the questions covering the same concept are considered as a single question. Thus the questions with the same concept have the same one-hot encoding.}. In practice, $x_i$ is encoded using a one-hot vector $\textbf{x}_i$ of length $2M$. If $a_i=0$, we concatenate $\textbf{e}_i$ with a zero vector $\textbf{z}$ of length $M$ to form $\textbf{x}_i$; otherwise, we concatenate $\textbf{z}$ before $\textbf{e}_i$~\cite{piech2015deep,zhang2017dynamic,huang2019ekt}. The encoding process can be summarized as follow:
\begin{equation}
  \label{equ:encode_onehot}
  \textbf{x}_i = \left\{
  \begin{aligned}
    &\lbrack\textbf{e}_i \oplus \textbf{z}\rbrack\quad  & \mathrm{if} \ a_i = 0, \\ 
    &\lbrack\textbf{z} \oplus \textbf{e}_i\rbrack\quad  & \mathrm{if} \ a_i = 1,
  \end{aligned}
  \right.
\end{equation}
where $\oplus$ represents the concatenation operation.

\begin{figure*}[t]
  \centering
  \includegraphics[width=1.0\linewidth]{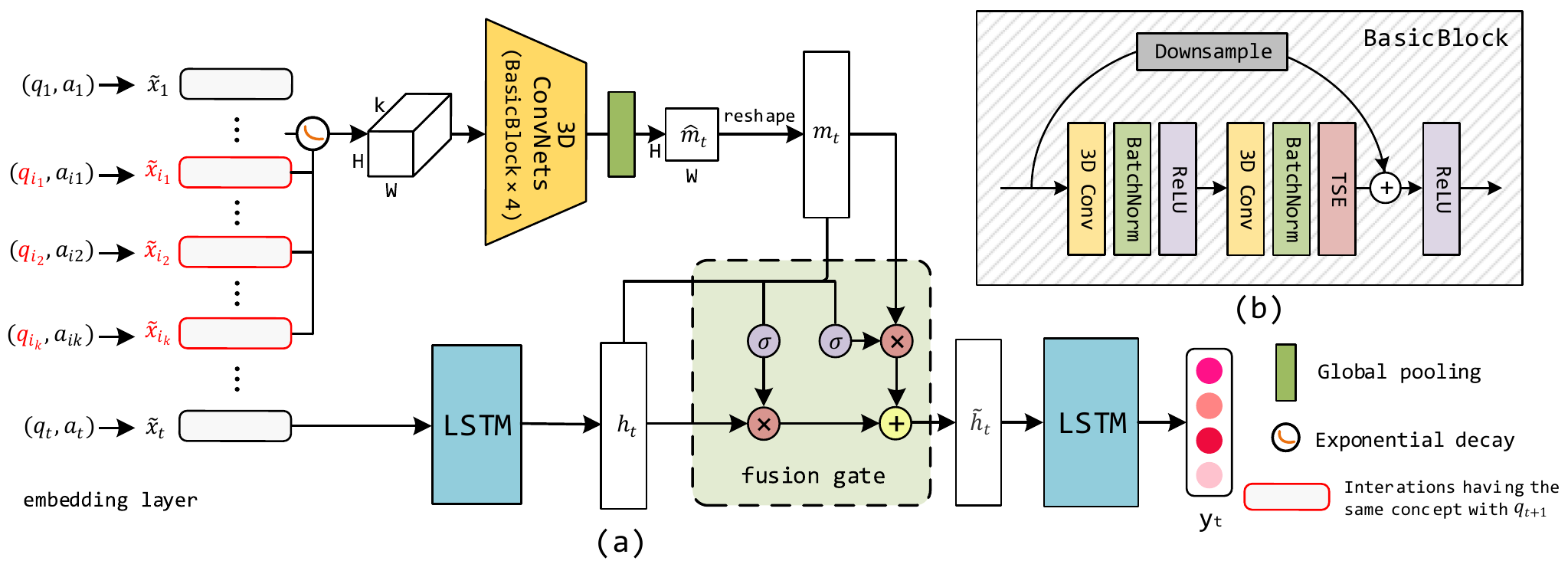}
  \caption{(a) The overall architecture of CAKT. The red rounded rectangles represent the embeddings of the interactions that apply the same knowledge concept with that of $q_{t+1}$. The 3D-ConvNets and the left LSTM layer extract the features $\mathbf{m}_t$ pertaining to the latent knowledge state on the concept covered by $q_{t+1}$ and the overall latent knowledge state $\mathbf{h}_t$, respectively. The component to the left of the tensor represents the exponential decay function. The 3D-ConvNets module consists of four BasicBlocks depicted in (b). The fusion gate adaptively integrates the two features. Finally, the right LSTM layer transforms the fused feature and outputs the integrated knowledge state $\mathbf{y}_t$. (b) Each BasicBlock consists of two 3D convolutional layers, two batch normalization layers, two ReLU layers and one TSE layer. It also employs a residual connection between the input to the block and the output of the TSE layer.}
  \label{fig:model}
\end{figure*}

\section{The CAKT Model} \label{sec:models}
\subsection{Model Overview} \label{sec:overview}
Figure~\ref{fig:model}(a) illustrates the architecture of CAKT, which consists of two sub-modules 3D-ConvNets and LSTM networks. The LSTM module consists of two LSTM layers (left and right), and the 3D-ConvNets module functions between the two LSTM layers.

In the original input sequence $\mathcal{X} = \{\mathbf{x}_1, \mathbf{x}_2, \dots, \mathbf{x}_t\}$, each interaction $\mathbf{x}_i$ is a one-hot vector of length $2M$, as described in Section~\ref{sec:input}. We employ an embedding layer to convert each $\mathbf{x}_i$ into a dense embedding $\tilde{\mathbf{x}}_i$ with dimension $d_e$, as depicted in the leftmost part of the architecture. The reasons are twofold. First, for datasets with a large number of unique knowledge concepts, such as Statics2011 described in section~\ref{sec:datasets}, a one-hot encoding can quickly become impractically large~\cite{piech2015deep}. Second, compared with one-hot encodings, it is much easier for convolutional neural networks to find interesting patterns on the denser representations. Then we use a classic LSTM layer as in DKT to learn from the entire sequence of embedded interactions $\mathcal{\tilde{X}} = \{\tilde{\mathbf{x}}_1, \tilde{\mathbf{x}}_2, \dots, \tilde{\mathbf{x}}_t\}$, and output a latent representation $\mathbf{h}_t$ with length $d_h$ at each time step $t$, i.e., $\mathbf{h}_t$ captures the student's historical performance on all the questions until $t$. This is depicted in the bottom-middle of the architecture. Note that we just draw the LSTM unit at time $t$ and omit all previous units for the sake of simplicity.

To incorporate the learning curve theory into the model, when predicting a student's response to $q_{t+1}$ with knowledge concept $c$, we additionally investigate how she has performed on the $k$ most recent questions before time $t$+$1$ covering the same concept $c$. The structures are depicted in the top part of the architecture. Specifically, we pick the $k$ most recent embedded interactions $\tilde{\mathbf{x}}_{i_1}, \tilde{\mathbf{x}}_{i_2},..., \tilde{\mathbf{x}}_{i_k}$ that contain $c$, as shown by the red rounded rectangles in the left part of the architecture. If there are less than $k$ interactions with the concept before time $t$+$1$, we use the all-zero embeddings to compensate. Before reshaping the embeddings to form a tensor, we take into account the forgetting curve hypothesis~\cite{ebbinghaus2013memory}, which states that the human memory retention declines over time, and thus give bias to each embedding according to its time gap to $t$+$1$. The simplest way of simulating the phenomenon is to use an exponential decay function~\cite{wozniak1995two}. Therefore, we propose the following equation to transform the values in the $k$ embeddings:
\begin{equation}
  \tilde{\mathbf{x}}_i = \exp(-\frac{\Delta{t_i}}{\theta}) \times \tilde{\mathbf{x}}_i,
  \label{equ:exponential_decay}
\end{equation}
where $\Delta{t_i}$ is the time gap between interaction $\tilde{\mathbf{x}}_i$ and time $t$+$1$, $\theta$ is a learnable parameter which controls the rate of decay. As such the interactions in the long past have small impact on the current knowledge state. Then we reshape each of the $k$ embeddings into a matrix (feature map) with shape $H\times W$, where $H\times W=d_e$. In CAKT, we set $H=W$, but in practice one may set it to any shape as long as the equation holds. We stack the $k$ matrices in their original chronological order and form a three-dimensional tensor of shape $k \times H \times W$. Then we feed the tensor into a 3D-ConvNets module which consists of four BasicBlocks. The architecture of BasicBlock is depicted in Figure~\ref{fig:model}(b). The 3D-ConvNets output a tensor with the same shape as the input tensor, followed by a global average pooling layer to squash the output tensor in the time dimension into a matrix of shape $H\times W$. Finally, the squashed matrix is stretched into the latent vector $\mathbf{m}_t$ of size $d_e$. Without loss of generality, we set $d_e = d_h$ so that $\mathbf{m}_t$ and $\mathbf{h}_t$ have the same length. The depth $k$ of the tensor and the embedding size $d_e$ ($d_h$) are the two hyper-parameters to be adjusted. 

Now we obtain two hidden state vectors $\mathbf{h}_t$ and $\mathbf{m}_t$, representing the student's overall latent knowledge state and the latent knowledge state on concept $c$ covered by $q_{t+1}$, respectively. In order to integrate the two features, we borrow the idea of the threshold mechanism in LSTM\textbackslash GRU and propose a fusion gate to adaptively fuse them. The fusion gate outputs the hidden state $\tilde{\mathbf{h}}_t$ at each step $t$, which is fed as input to the right LSTM layer in the figure. The right LSTM layer finally outputs a prediction vector $\mathbf{y}_t$ with length $M$, each element of which represents the probability that the student has mastered the corresponding knowledge concept at time $t$. Similarly to the left LSTM layer, we omit in the figure all LSTM units except the one of step $t$ for the sake of simplicity. Here one may also use a fully-connected layer instead, but we will show this second LSTM layer yields better results. The predicted response to the next question $q_{t+1}$ can be directly read from the element in $\mathbf{y}_t$ corresponding to the concept covered by $q_{t+1}$.

\subsection{The 3D-ConvNets Module}
\label{sec:3Dconv}
The 3D-ConvNets module takes as input the three-dimensional tensor with shape $k \times H \times W$, which wraps the information in the $k$ most recent interactions with the same concept covered by $q_{t+1}$. We design a block named \emph{BasicBlock} as shown in Figure~\ref{fig:model}(b), and stack four BasicBlocks to form the 3D-ConvNets. 

A BasicBlock consists of two three-dimensional convolutional layers, each of which is followed by a batch normalization layer and a ReLU layer. In addition, we use a residual connection to sum the input to BasicBlock and the output before the second ReLU layer. To ensure the input and output have the same shape, we use a three-dimensional filter with size $1 \times 1 \times 1$ to convolve the input, which is depicted as the Downsample component in Figure~\ref{fig:model}(b). The residual connections force the BasicBlocks to learn the residual features~\cite{he2016deep} from the input tensors and facilitate the network optimization. 

\begin{figure}[t]
  \centering
  \includegraphics[width=1\linewidth]{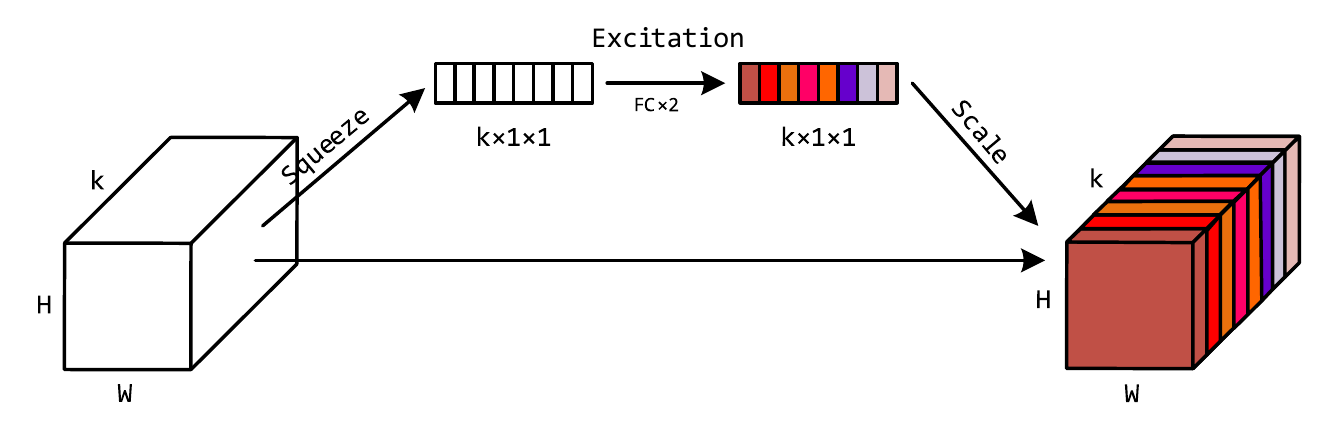}
  \caption{The architecture of the Timely-Squeeze-and-Excitation (TSE) layer. The squeeze stage uses global pooling to transform the input tensor with shape $k \times H \times W$ into a vector with length $k$. The excitation stage uses two fully-connected layers to transform the $k$ entries in the vector into values between 0 and 1. The scale stage uses these $k$ values as importance weights to multiply with corresponding feature matrices in the input tensor to TSE.}
  \label{fig:TSE}
\end{figure}
In addition to the exponential decay function applied to the input embeddings, we want the 3D-ConvNets to further adaptively learn the importance of the feature maps at different time steps. Inspired by the 
Squeeze-and-Excitation Networks~\cite{hu2018squeeze}, we design a Timely-Squeeze-and-Excitation (TSE) layer and put it right after the second batch normalization layer in each BasicBlock. The architecture of TSE is shown in Figure~\ref{fig:TSE}. For the sake of simplicity, we ignore the batch and the channel dimension of the tensors. The input tensor to TSE with shape $k \times H \times W$ is squeezed into a vector of length $k$ using global pooling. Then we employ two fully-connected layers for 
excitation, where the first layer transforms the squeezed vector to length $\frac{k}{2}$ and uses ReLU for activation, and the second layer converts the vector back to length $k$ and uses the Sigmoid function for activation. Each entry of the excited vector has value between 0 and 1, which indicates the importance of the corresponding feature matrix in the input tensor to TSE. Finally, we scale the input tensor to TSE by multiplying each feature matrix with its importance weight, and form a new tensor with shape $k \times H \times W$, as depicted in the rightmost part of Figure~\ref{fig:TSE}.

In total, the BasicBlock takes as input a three-dimensional tensor with shape $k \times H \times W$ and outputs a tensor with the same shape. We stack four BasicBlocks to form the 3D-ConvNets module. In the experiments we find that using small filters yields better results. Hence, following the principles in VGG~\cite{simonyan2014very} and FCN~\cite{long2015fully}, we use filters with size $3 \times 3 \times 3$ in all the BasicBlocks and discard the pooling layers. We also use stride size with 1 and perform one zero-padding to ensure the tensor size unchanged. The filter size and the input\textbackslash output shape in each BasicBlock are presented in Table~\ref{tab:conv_params}.
\begin{table}
  \renewcommand\arraystretch{1.1}
  \newcommand{\tabincell}[2]{\begin{tabular}{@{}#1@{}}#2\end{tabular}}
  \caption{The filter size and the input\textbackslash output shapes in the BasicBlocks.}
  \label{tab:conv_params}
  \setlength{\tabcolsep}{1mm}{
  \begin{tabular}{ccccc}
    \toprule
    layer name & output size & filter size & in channel & out channel\\
    \midrule
    BasicBlock1 & $k \times \mathrm{H} \times \mathrm{W}$ & \tabincell{c}{$3 \times 3 \times 3$\\ $3 \times 3 \times 3$} & \tabincell{c}{1\\ 4} & \tabincell{c}{4\\ 4} \\
    \midrule
    BasicBlock2 & $k \times \mathrm{H} \times \mathrm{W}$ & \tabincell{c}{$3 \times 3 \times 3$\\ $3 \times 3 \times 3$} & \tabincell{c}{4\\ 8} & \tabincell{c}{8\\ 8} \\
    \midrule
    BasicBlock3 & $k \times \mathrm{H} \times \mathrm{W}$ & \tabincell{c}{$3 \times 3 \times 3$\\ $3 \times 3 \times 3$} & \tabincell{c}{8\\ 4} & \tabincell{c}{4\\ 4} \\
    \midrule
    BasicBlock4 & $k \times \mathrm{H} \times \mathrm{W}$ & \tabincell{c}{$3 \times 3 \times 3$\\ $3 \times 3 \times 3$} & \tabincell{c}{4\\ 1} & \tabincell{c}{1\\ 1} \\
  \bottomrule
\end{tabular}}
\end{table}

After the convolution operations, we use a global average pooling layer to squash the output tensor on the time dimension into a matrix $\hat{\mathbf{m}}_t$ of shape $H \times W$, which can be formulated as follows:
\begin{equation}
  \hat{\mathbf{m}}_t = \frac{1}{k} \sum_{i=1}^{k} \mathbf{m}^i,
\end{equation}
where $k$ is the depth of the output tensor and $\mathbf{m}^i$ is the $i^{th}$ feature map in the output tensor. Then $\hat{\mathbf{m}}_t$ is further stretched into a hidden vector $\mathbf{m}_t$ of size $d_e$, which represents the latent knowledge state on the concept covered by $q_{t+1}$.

\subsection{Adaptive Feature Fusion}
In order to fuse $\mathbf{h}_t$ and $\mathbf{m}_t$, we borrow the idea from the threshold mechanism in LSTM\textbackslash GRU and propose a fusion gate to adaptively learn the weights of the features. The weights control how much information of the two latent features should be preserved. The process can be formulated as follows:
  \begin{align}
  \mathbf{z}^1_t &= \sigma(\left[\mathbf{m}_t \oplus \mathbf{h}_t \right]\mathbf{W}^1_z + \mathbf{b}^1_z) \label{equ:fusion_gate1}\\
  \mathbf{z}^2_t &= \sigma(\left[\mathbf{m}_t \oplus \mathbf{h}_t \right]\mathbf{W}^2_z + \mathbf{b}^2_z) \label{equ:fusion_gate2}\\
  \tilde{\mathbf{h}}_t &= \mathbf{z}^1_t \odot \mathbf{m}_t + \mathbf{z}^2_t \odot \mathbf{h_t} \label{equ:fuse}
  \end{align}
where $\oplus$ represents concatenating two vectors, $\mathbf{W}^1_z$ and $\mathbf{W}^2_z$ are two weight matrices with shape $2 d_e \times d_e$, $\mathbf{b}^1_z$ and $\mathbf{b}^2_z$ are two bias vectors with length $d_e$, $\sigma$ represents the Sigmoid function and $\odot$ represents the Hadamard product between two vectors. In particular, we concatenate $\mathbf{m}_t$ with $\mathbf{h}_t$ into a vector with length $2d_e$, and then use two fully-connected layers with Sigmoid activation to transform the vector into two gates $\mathbf{z}^1_t$ and $\mathbf{z}^2_t$, which are shown in Equation~\ref{equ:fusion_gate1} and Equation~\ref{equ:fusion_gate2}. The two gates control the information preserved in $\mathbf{m}_t$ and $\mathbf{h}_t$, respectively. We can then fuse the two features into a single feature $\tilde{\mathbf{h}}_t$, as shown in equation~\ref{equ:fuse}.

Finally, we use an additional LSTM layer to transform $\tilde{\mathbf{h}}_t$ and obtain the predicted knowledge state vector $\mathbf{y}_t$ with length $M$, where each element represents the probability that the student has mastered the corresponding knowledge concept at time $t$.

\subsection{Objective Function}
The objective function is a binary cross-entropy loss function, calculated using the predicted probability $p_t$ that $q_t$ is correctly answered and the ground-truth response $a_t$, for all time step $t$. As we discussed in Section~\ref{sec:overview}, $p_t$ can be directly read from the element in $\mathbf{y}_{t-1}$ corresponding to the concept in $q_{t}$. The function can be formulated as:
\begin{equation}
  \mathcal{L} = - \sum_t(a_t\ \mathrm{log}\ p_t + (1-a_t) \mathrm{log}(1-p_t))
\end{equation}

\section{Experiments}\label{sec:experiment}
\subsection{Datasets}\label{sec:datasets}
We use five datasets to evaluate the CAKT model and compare it with existing models and its alternative architectures. The statistics of the datasets are shown in Table~\ref{tab:datasets}.
\begin{table}[h]
  \renewcommand\arraystretch{1.2}
  \newcommand{\tabincell}[2]{\begin{tabular}{@{}#1@{}}#2\end{tabular}}
  \caption{Statistics of the datasets.}
  \label{tab:datasets}
  \setlength{\tabcolsep}{1mm}{
  \begin{tabular}{c|ccc|c}
    \toprule
    \small{Dataset} & \tabincell{c}{\small{\#Questions}\\\small{(\#Concepts)}} & \small{\#Students} & \tabincell{c}{\small{\#Interactions}\\\small{(\#Exercises)}} & \tabincell{c}{\small{\#Interactions}\\\small{per student}}\\
    \midrule
    \small{ASSISTments2009} & 110 & 4,151 & 325,637 & 78\\
    \small{ASSISTments2015} & 100 & 19,840 & 683,801 & 34\\
    \small{ASSISTments2017} & 102 & 1,709 & 942,816 & 552\\
    \small{Statics2011} & 1,223 & 333 & 189,297 & 568\\
    \small{Synthetic-5} & 50 & 4,000 & 200,000 & 50\\
  \bottomrule
\end{tabular}}
\end{table}
\begin{itemize}
  \item $\textbf{ASSISTments2009}\footnote{ASSISTments2009: https://sites.google.com/site/assistmentsdata/home/assistment-2009-2010-data/skill-builder-data-2009-2010}:$ This dataset was gathered in the school year 2009-2010 from the ASSISTments education platform. We use the skill  builder data of ASSISTments2009, which consists of 110 distinct questions (knowledge concepts),  4,151 students and 325,637 exercise records  (student-question interactions).
  \item $\textbf{ASSISTments2015}\footnote{ASSISTments2015:https://sites.google.com/site/assistmentsdata/home/2015-assistments-skill-builder-data}:$ This dataset was collected in 2015. It is an updated variant to ASSISTments2009. It includes 100 distinct questions, 19,840 students and 683,801 exercise records. This dataset has the largest number of students, but the average number of exercise records per student (34) is the smallest among the five datasets.
  \item $\textbf{ASSISTments2017}\footnote{ASSISTments2017: https://sites.google.com/view/assistmentsdatamining}:$ This dataset was collected from the ASSISTments education platform in 2017. It includes 102 distinct questions, 1,709 students and 942,816 exercise records.
  \item $\textbf{Statics2011}\footnote{Statics2011: https://pslcdatashop.web.cmu.edu/DatasetInfo?datasetId=507}:$ This dataset was collected from a statistics course at Carnegie Mellon University in the fall of 2011. It contains 1,223 distinct questions, 333 students, and 189,297 exercise records.
  \item $\textbf{Synthetic-5}\footnote{Synthetic-5: https://github.com/chrispiech/DeepKnowledgeTracing/\\tree/master/data/synthetic}:$ This is a synthetic dataset generated by Piech et al. \cite{piech2015deep}. It contains 50 distinct questions, 4,000 virtual students and 200,000 exercise records. All the students answer the same 50 questions in the same order.
\end{itemize}

Among these datasets, Statics2011 and ASSISTments2017 have fewer interaction sequences, but the lengths of the sequences are typically long. So following the methods in related work~\cite{piech2015deep,zhang2017dynamic}, we conduct a fold operation on the two datasets. In particular, when the length of a sequence exceeds 200, we split the sequence into sub-sequences so that the length of each sub-sequence is less than or equal to 200.

\begin{table*}[t]
  \renewcommand\arraystretch{1.1}
  \renewcommand\arraycolsep{1}
  \newcommand{\tabincell}[2]{\begin{tabular}{@{}#1@{}}#2\end{tabular}}
  \caption{The test AUC results (\%) of all the models. The results of CAKT are in the last column and bolded, and all results that are better than CAKT are also bolded.}
  \label{tab:result}
  \begin{tabular}{c|cccccccccc}
  \toprule
  Dataset & DKT & DKVMN & SKVMN & SAKT & EKT-R & EKT-C & DKT-F & CKT & AKT-NR & CAKT\\
  
  \midrule
  ASSISTments2009 & 81.19 &	80.02 &	67.39 &	76.59 &	76.46 &	76.45 &	81.88 &	82.13 &	81.84 &	\textbf{82.37}\\
  \midrule
  ASSISTments2015 & 71.95  & 72.33  & 67.01  & 73.27  & 70.65  & 70.35  & 72.96  & \textbf{73.45}  & \textbf{73.43}  & \textbf{73.31}\\
  \midrule
  ASSISTments2017 & 64.47  & 68.53  & 56.95  & 64.85  & 60.25  & 61.56  & 73.48  & 72.16  & 72.06  & \textbf{73.68}\\
  \midrule
  Statics2011 & 79.00  & 80.42  & 78.41  & 81.43  & 75.65  & 77.73  & 82.76  & 82.38  & 82.74  & \textbf{82.78}\\
  \midrule
  Synthetic-5 & 81.03  & \textbf{82.60}  & 75.46  & \textbf{82.53}  & 78.50  & 75.20  & 82.26  & \textbf{82.85}  & \textbf{83.39}  & \textbf{82.28}\\
  \bottomrule
  \end{tabular}
\end{table*}

\subsection{Network Instances of CAKT}
The important hyperparameters for tuning are the depth of the input tensor $k$, the embedding size of the input interaction $d_e$ and the hidden state size of the first LSTM layer $d_h$. To facilitate the fusion of $\mathbf{m}_t$ and $\mathbf{h}_t$, we set $d_e=d_h$. The reshaped interaction matrix satisfies $H \times W = d_e$, where $H$ and $W$ are the height and width. Without loss of generality, we set $H=W$ and therefore $d_e$ must be the square of an integer value. One may set $H$ not equal to $W$ as long as $H \times W = d_e$ holds. As such we just need to tune $k$ and $d_e$ in the training phase. We also perform a sensitivity analysis on $k$ and $d_e$ in Section~\ref{sec:sensitive}.

The 3D-ConvNets module is constructed by stacking four BasicBlocks, each of which contains a Conv-BN-ReLU layer and a Conv-BN-TSE-ReLU layer with a residual link. We set the filter size of the convolutional layers to $3 \times 3 \times 3$ as discussed in Section~\ref{sec:3Dconv}, and vary their channel sizes in the forward pass. The sizes of the BN and ReLU layers are decided by the convolultional layers. The left and right LSTM layer use the same hidden size. 

All source code is available at https://github.com/Badstu/CAKT.

\subsection{The Comparative Models}
We compare CAKT with eight state-of-the-art deep models for the knowledge tracing task, 
namely DKT~\cite{piech2015deep}, DKVMN~\cite{zhang2017dynamic}, SKVMN~\cite{abdelrahman2019knowledge}, SAKT~\cite{pandey2019self}, EKT~\cite{huang2019ekt}, CKT~\cite{shen2020convolutional}, DKT-F (DKT+forgetting)~\cite{nagatani2019augmenting} and AKT~\cite{ghosh2020context}. Among the models, EKT has two variants, which employ the Markov chain and attention mechanism, respectively. We choose the variant using attentions since the authors report better results with it. One issue is that EKT requires the text information of questions as input. Since the dataset with text information is not available, we slightly modify the input features of EKT and obtain two variants for comparison. The first one uses a fixed randomized embedding to represent the text information for each distinct question, which is referred to as EKT-R. The second one replaces the randomized text information embedding with the knowledge concept embedding at each time step, which is referred to as EKT-C. The AKT model has also two variants in~\cite{ghosh2020context}, namely AKT-R and AKT-NR. AKT-R requires both knowledge ID and question ID as input, whereas AKT-NR only requires knowledge ID as in the case of other models. For fair comparison, we choose the AKT-NR model since all other models do not use the question ID information. In addition, the datasets except ASSISTments2009 and ASSISTments2017 do not contain the information of question ID~\cite{ghosh2020context}.

We reimplement SKVMN and DKT-F since we have not found the released source code. We use the source code on GitHub for DKT~\footnote{https://github.com/mmkhajah/dkt}, DKVMN~\footnote{https://github.com/jennyzhang0215/DKVMN}, EKT~\footnote{https://github.com/bigdata-ustc/ekt}, CKT~\footnote{https://github.com/shshen-closer/Convolutional-Knowledge-Tracings} and AKT~\footnote{https://github.com/arghosh/AKT}. We obtain the code for SAKT from the original authors.

\subsection{Model Training and the Evaluation Metric}
We implement CAKT using Pytorch 1.6 and train it on an NVIDIA Tesla-V100 card with 16GB memory. We use the Adam optimizer to optimize the network parameters. We set the L2 regularization term to 1e-5 and the initial learning rate is 0.001, with decay of 0.3 every 5 epochs.

Similarly to the settings in~\cite{piech2015deep,zhang2017dynamic,abdelrahman2019knowledge,ghosh2020context}, we use $20\%$ of the interaction sequences to form a testing set for each dataset, and split the remaining sequences into five folds for cross validation. For each dataset, the hyperparameters of each model are determined when it has the best average performance on the validation sets. Then we report the corresponding average results on the testing set for each model on each dataset. The evaluation metric on the testing set is the Area Under the ROC Curve, referred to as AUC~\cite{Belohradsky2011}, which is commonly used to evaluate the performance of knowledge tracing models~\cite{piech2015deep,zhang2017dynamic,abdelrahman2019knowledge,ghosh2020context}. When AUC=0.5, it means that the prediction makes no difference from a random guess. The higher value the AUC, the better the prediction performance of the model.

\subsection{The Results of Comparative Evaluation}
\subsubsection{\textbf{Main Results}} The main comparative results are reported in Table~\ref{tab:result}. We use the bold font for the results of CAKT and some comparable results of existing models. We observe that our CAKT model performs the best among the existing models on three real-life datasets (ASSISTments2009, ASSISTments2017 and Statics2011). On ASSISTments2015 and the synthetic dataset Synthetic-5, CAKT also produces very comparable results with that of the state-of-the-art methods, such as CKT and AKT-NR. The overall results prove the usefulness of incorporating the learning curve theory when designing a deep model for knowledge tracing.

By analyzing ASSISTments2015 and Synthetic-5, we notice that the two datasets have relatively short average interaction sequences (34 and 50, respectively) per student. Unlike the other three datatsets, students don't practice many times on the same knowledge concept. As a result, the 3D-ConvNets module in CAKT may not learn enough information about a student's past experience on applying the knowledge concept covered by the question to be answered at each step, which in turn leads to the degeneration of the prediction performance of CAKT on the two datasets. This indeed implies the importance of modeling the learning curve theory, if only there are sufficient number of past trials on the same knowledge concept.

It is worth noting that CAKT greatly improves the performance of DKT on all datasets. Remember that CAKT retains the LSTM structure used in DKT to learn from the entire sequence in the past. Therefore such big performance gain over DKT is mainly due to the explicit modeling of the learning curve theory. While this proves the importance of the modeling, more advanced structures may be used to replace LSTM and further boost the KT performance.

\subsubsection{\textbf{Convergence Rate}} 
\begin{figure}[t]
  \centering
  \includegraphics[width=1\linewidth]{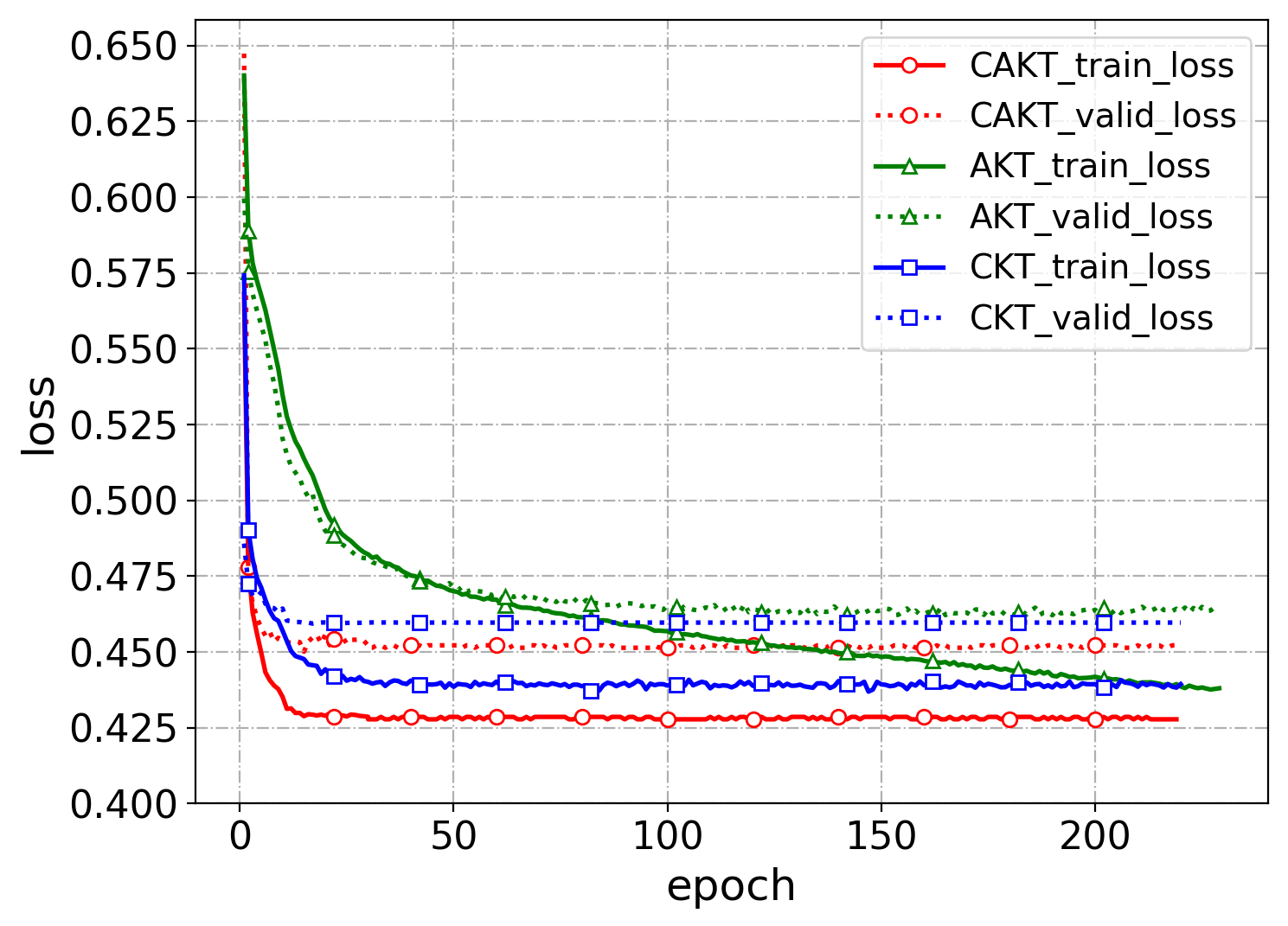}
  \caption{Convergence rate. The horizontal axis represents the number of epochs, and the vertical axis represents the loss values.}
  \label{fig:convergence}
\end{figure}

In Table~\ref{tab:result} we observe CKT and AKT-NR have the overall closest performance to that of CAKT. In this section, we further compare the convergence rates of the three models and show the results on the ASSISTments2009 dataset in Figure~\ref{fig:convergence}. The results on the other datasets are similar and thus omitted. However, they are available upon request. We use the hyperparameters tuned in the main results for each model. In particular, CAKT uses 6 for $k$ and 289 for $d_e$ (thus $H=W=17$). The initial learning rate is 0.001 with decay of 0.3 every 5 epochs, and the L2 regularization term is 1e-5 in CAKT. Both CKT and AKT-NR use learning rate 0.001. AKT-NR sets the L2 regularization to 1e-5 and CKT does not have a regularization item. In the figure, we observe that our CAKT model converges within 10 epochs, which is the fastest among the three models. CKT converges in about 20 epochs. AKT-NR does not converge even after training for 200 epochs and becomes slightly over-fitting after 100 epochs since the validation loss does not decrease anymore. In the original paper of AKT-NR~\cite{ghosh2020context}, the authors perform early stopping to avoid over-fitting. In summary, CAKT converges much faster than CKT and AKT-NR, while achieving the overall better performance on AUC.

\subsection{Sensitivity Analysis}
\label{sec:sensitive}
In this part, we explore the influence of three hyper-parameters in our CAKT model. We conduct sensitivity analysis on the number of recent interactions $k$ on the same knowledge concept, the batch size $b$, and the height $H$ of the input tensor to the 3D ConvNets module. For each experiment, we fix two hyperparameters and adjust the remaining one.

The first experiment focuses on the hyperparameter $k$, that is, how many past interactions on the same knowledge concept are needed for capture the experience. We fix $b = 32$ and $H = 17$, and vary $k$ from 4 to 14 with increments 2. The results on each dataset are reported in Table~\ref{tab:sen1}. We observe that the performances are relatively stable and a moderate $k$ value can already bring the best results. This is rationale because the most recent interactions have the highest impact on the current knowledge state, according to the forgetting curve hypothesis~\cite{ebbinghaus2013memory}. On the other hand, a moderate $k$ indicates that CAKT just needs a small 3D ConvNets module, which reduces the complexity of the model.
\begin{table}
  \renewcommand\arraystretch{1.1}
  \renewcommand\arraycolsep{1}
  \newcommand{\tabincell}[2]{\begin{tabular}{@{}#1@{}}#2\end{tabular}}
  \caption{The AUC results (\%) for varying $k$. The best AUC in each column is bolded.}
  \label{tab:sen1}
  \begin{tabular}{c|c|c|c|c|c|c}
    \toprule
  \multirow{2}{*}{b\&H} & \multirow{2}{*}{k} & \multirow{2}{*}{\shortstack{ASSIST\\2009}} & \multirow{2}{*}{\shortstack{ASSIST\\2015}} & \multirow{2}{*}{\shortstack{ASSIST\\2017}} & \multirow{2}{*}{Statics} & \multirow{2}{*}{Synthetic}\\
  ~ & ~ & ~ & ~ & ~ & ~ & ~\\
    \midrule
    \multirow{8}{*}{\tabincell{c}{b=32\\ H=17}} & k=4 & 82.23 & 73.27 & 71.95 & 81.48 & 82.12\\
    ~ & k=6 & \textbf{82.37} & 73.21 & 72.94 & 80.99 & 82.02\\
    ~ & k=8 & 82.08 & 73.23 & 72.14 & 81.21 & \textbf{82.28}\\
    ~ & k=10 & 82.02 & \textbf{73.29} & 71.84 & \textbf{82.49} & 82.13\\
    ~ & k=12 & 81.99 & 72.97 & 73.18 & 82.28 & 81.92\\
    ~ & k=14 & 81.98 & 73.28 & \textbf{73.32} & 81.40 & 81.95\\
  \bottomrule
  \end{tabular}
\end{table}

The second experiment focuses on the batch size $b$. We fix $k = 6$ and $H = 17$, and vary the value of $b$. The results are reported in Table~\ref{tab:sen2}. We observe that overall CAKT just needs a small batch size to achieve good results. On ASSISTments2009, ASSISTments2017 and Statics, the best results are obtained when the batch sizes are set to 32, 8 and 8, respectively.
\begin{table}
  \renewcommand\arraystretch{1.1}
  \renewcommand\arraycolsep{1}
  \newcommand{\tabincell}[2]{\begin{tabular}{@{}#1@{}}#2\end{tabular}}
  \caption{The AUC results (\%) for varying $b$. The best AUC in each column is bolded.}
  \label{tab:sen2}
  \begin{tabular}{c|c|c|c|c|c|c}
    \toprule
  \multirow{2}{*}{k\&H} & \multirow{2}{*}{b} & \multirow{2}{*}{\shortstack{ASSIST\\2009}} & \multirow{2}{*}{\shortstack{ASSIST\\2015}} & \multirow{2}{*}{\shortstack{ASSIST\\2017}} & \multirow{2}{*}{Statics} & \multirow{2}{*}{Synthetic}\\
  ~ & ~ & ~ & ~ & ~ & ~ & ~\\
    \midrule
    \multirow{7}{*}{\tabincell{c}{k=6\\ H=17}} & b=8 & 81.84 & 73.07 & \textbf{73.66} & \textbf{82.64} & 81.77\\
    ~ & b=16 & 81.98 & 73.13 & 73.38 & 82.08 & 82.06\\
    ~ & b=32 & \textbf{82.37} & 73.21 & 72.94 & 80.99 & 82.02\\
    ~ & b=48 & 82.07 & 73.19 & 73.31 & 81.07 & 82.09\\
    ~ & b=64 & 82.23 & \textbf{73.30} & 72.78 & 81.28 & \textbf{82.17}\\
    ~ & b=80 & 82.09 & 73.26 & 73.03 & 82.13 & 81.89\\
    ~ & b=96 & 81.92 & 73.23 & 72.26 & 80.77 & 82.02\\
  \bottomrule
  \end{tabular}
\end{table} 

The third experiment focuses on $H$. We fix $k = 6$ and $b = 32$, and vary $H$ from 11 to 19 with increments 2. In CAKT we set $H = W$ and $H\times W=d_e=d_h$, thus $H$ decides the size of the feature map, the input embedding size and the hidden vector size of the two sub-modules. The results are reported in Table~\ref{tab:sen3}. We observe that overall a bigger $H$ brings better results, since big feature maps or long vectors usually contain more information. However, the values of $H$ are still moderate, making both the sizes of the input tensor and embedding small.
\begin{table}
  \renewcommand\arraystretch{1.1}
  \renewcommand\arraycolsep{1}
  \newcommand{\tabincell}[2]{\begin{tabular}{@{}#1@{}}#2\end{tabular}}
  \caption{The AUC results (\%) for varying $H$. The best AUC in each column is bolded.}
  \label{tab:sen3}
  \begin{tabular}{c|c|c|c|c|c|c}
  \toprule
  \multirow{2}{*}{k\&b} & \multirow{2}{*}{H} & \multirow{2}{*}{\shortstack{ASSIST\\2009}} & \multirow{2}{*}{\shortstack{ASSIST\\2015}} & \multirow{2}{*}{\shortstack{ASSIST\\2017}} & \multirow{2}{*}{Statics} & \multirow{2}{*}{Synthetic}\\
  ~ & ~ & ~ & ~ & ~ & ~\\
  \midrule
  \multirow{5}{*}{\tabincell{c}{k=6\\ b=32}} & H=11 & 81.9 & 73.22 & 72.17 & 82.14 & 81.64\\
  ~ & H=13 & 81.97 & 73.21 & 72.68 & 81.68 & \textbf{82.16}\\
  ~ & H=15 & 82.23 & 73.23 & 72.96 & 82.45 & 82.11\\
  ~ & H=17 & \textbf{82.37} & 73.21 & 72.94 & 80.99 & 82.02\\
  ~ & H=19 & 81.95 & \textbf{73.29} & \textbf{73.65} & \textbf{82.78} & 82.06\\
  \bottomrule
  \end{tabular}
\end{table}

\subsection{Ablation Study}
We conduct eight ablation experiments to show the effectiveness of the components designed in CAKT. The first four experiments pertain to using alternative structures to capture the latent knowledge state $\mathbf{m}_t$ to simulate the learning curve phenomenon. We replace the 3D-ConvNets with an LSTM layer, a fully-connected layer, a 2D convolutional layer and a self-attention layer, respectively, and obtain four alternative models, namely, $\mathbf{LSTM\_LC}$, $\mathbf{FC\_LC}$, $\mathbf{C2D\_LC}$, $\mathbf{SA\_LC}$. The other four experiments examine the usefulness of other components in CAKT. In particular, we directly stack the $k$ recent matrices pertaining to the same knowledge concept without considering the decayed impact and call this model $\mathbf{NO\_EXP\_DECAY}$. We replace the global pooling layer with a fully-connected layer to compress the output tensor to $\hat{\mathbf{m}}_t$ and call this model $\mathbf{FC\_POOLING}$. We replace the right LSTM layer with a fully-connected layer to output $\mathbf{y}_t$ and call this model $\mathbf{FC\_OUTPUT}$. Finally, we replace the fusion gate with the mean of the two latent features $\mathbf{m}_t$ and $\mathbf{h}_t$ and call this model $\mathbf{MEAN\_FUSION}$. The results are reported in Table~\ref{tab:ablation}. We observe that the original CAKT model has the highest AUC values on all the datasets, which proves the design choices of all the components. In particular, when we replace the 3D-ConvNets module with other common structures in the first four ablation models, the performance decreases drastically on all datasets. This may verify our conjecture in Section~\ref{sec:introduction}, that the 3D-ConvNets can capture latent knowledge state of each step differently from that captured by sequence structures, as well as the evolution of the knowledge state.
\begin{table}
  \renewcommand\arraystretch{1.2}
  \renewcommand\arraycolsep{1}
  \newcommand{\tabincell}[2]{\begin{tabular}{@{}#1@{}}#2\end{tabular}}
  \caption{The AUC results (\%) of ablation study.}
  \label{tab:ablation}
  \setlength{\tabcolsep}{1.3mm}{
  \begin{tabular}{c|c|c|c|c|c}
      \toprule
  \multirow{2}{*}{Ablation} & \multirow{2}{*}{\shortstack{ASSIST\\2009}} & \multirow{2}{*}{\shortstack{ASSIST\\2015}} & \multirow{2}{*}{\shortstack{ASSIST\\2017}} & \multirow{2}{*}{Statics} & \multirow{2}{*}{Synthetic}\\
  ~ & ~ & ~ & ~ & ~ & ~\\
      \midrule
      LSTM\_LC & 80.57 & 71.63 & 71.82 & 79.23 & 79.91\\
      FC\_LC & 81.01 & 71.73 & 72.22 & 79.55 & 80.01\\
      C2D\_LC & 81.17 & 71.62 & 73.04 & 79.64 & 79.72\\
      SA\_LC & 80.46 & 71.56 & 72.32 & 79.56 & 81.39\\
      NO\_EXP\_DECAY & 82.09 & 73.28 & 72.65 & 79.57 & 82.16\\
      FC\_POOLING & 82.12 & 73.16 & 72.98 & 80.53 & 82.19\\
      FC\_OUTPUT & 81.95 & 73.09 & 71.31 & 80.44 & 81.71\\
      MEAN\_FUSION & 82.14 & 73.13 & 72.44 & 79.73 & 82.11\\
      ORIG\_CAKT & \textbf{82.37} & \textbf{73.31} & \textbf{73.68} & \textbf{82.78} & \textbf{82.28}\\
  \bottomrule
  \end{tabular}}
\end{table}

\section{Conclusion}\label{sec:conclusion}
In this paper, we propose a novel model which is called \emph{Convolution-augmented Knowledge Tracing} (CAKT) for the knowledge tracing task. We leverage three-dimensional convolutional neural networks to explicitly model the learning curve theory and obtain a student's latent knowledge state on the concept covered by the question to be answered. We also use the classic LSTM networks to learn the student's overall latent knowledge state. We then design an fusion gate to fuse the two latent features, and use the fused feature to predict the student's response to the next question. As such, when predicting a student's response to the next question, we collectively consider her recent experience on applying the concept covered by the question and her overall experience on all the knowledge concepts. Extensive experiments prove that our CAKT model outperforms existing models and its own variants. 

While the current work proves the importance of explicit modeling of the learning curve theory, in future we would investigate other structures to further boost the KT performance. An interesting question is what the 3D-ConvNets and the LSTM layer have learnt from the interaction sequences, respectively. We thus leave this for future work.


\bibliographystyle{ACM-Reference-Format}
\bibliography{CAKT}

\end{document}